\documentclass[accepted]{article}

\usepackage{microtype}
\usepackage{graphicx}
\usepackage{subfigure}
\usepackage{booktabs} 

\usepackage{tcolorbox}
\usepackage[utf8]{inputenc} 
\usepackage[T1]{fontenc}    
\usepackage{url}            
\usepackage{amsfonts}       
\usepackage{nicefrac}       
\usepackage{microtype}      
\usepackage{amssymb,amsmath,mathtools}
\usepackage{bm}
\usepackage{xcolor}
\usepackage[normalem]{ulem}
\newcommand{\stkout}[1]{\ifmmode\text{\sout{\ensuremath{#1}}}\else\sout{#1}\fi}
\usepackage{tikz}
\usetikzlibrary{shapes,arrows,calc,positioning,scopes}
\usepackage{enumitem}

 \definecolor{ml1}{rgb}{    0,    0.4470,    0.7410}
 \definecolor{ml2}{rgb}{     0.8500   , 0.3250   , 0.0980}
 \definecolor{ml3}{rgb}{     0.9290   , 0.6940  ,  0.1250}
 \definecolor{ml4}{rgb}{     0.4940  ,  0.1840  ,  0.5560}
  \definecolor{ml5}{rgb}{    0.4660  ,  0.6740  ,  0.1880}
  \definecolor{ml6}{rgb}{    0.3010  ,  0.7450  ,  0.9330}
  \definecolor{ml7}{rgb}{    0.6350  ,  0.0780 ,   0.1840}

\DeclareMathAlphabet{\mathbfsf}{\encodingdefault}{\sfdefault}{bx}{n}
\newcommand{\p}{\mathbfsf}
\newcommand{\s}{\mathbfsf}

\newcommand{\ben}{\begin{equation*}}
\newcommand{\een}{\end{equation*}}
\newcommand{\be}{\begin{equation}}
\newcommand{\ee}{\end{equation}}

\DeclareMathAlphabet{\mathbfsf}{\encodingdefault}{\sfdefault}{bx}{n}

\renewcommand{\Re}{\mathbb R}

\renewcommand{\exp}[1]{{\bf e}^{#1}}

\usepackage{hyperref}

\usepackage{icml2018}

\icmltitlerunning{Robust and Scalable Models of Microbiome Dynamics}

\begin{document}

\twocolumn[
\icmltitle{Robust and Scalable Models of Microbiome Dynamics}



\icmlsetsymbol{equal}{*}

\begin{icmlauthorlist}
\icmlauthor{Travis E.~Gibson}{HMS}
\icmlauthor{Georg K.~Gerber}{HMS}

\end{icmlauthorlist}

\icmlaffiliation{HMS}{Massachusetts Host-Microbiome Center, Brigham and Women's Hospital, Harvard Medical School, Boston, MA, USA}

\icmlcorrespondingauthor{TE Gibson}{tgibson@mit.edu}
\icmlcorrespondingauthor{GK Gerber}{ggerber@bwh.harvard.edu}

\icmlkeywords{Machine Learning, ICML}

\vskip 0.3in
]



\printAffiliationsAndNotice{}  

\begin{abstract}
Microbes are everywhere, including in and on our bodies, and have been shown to play key roles in a variety of prevalent human diseases. Consequently, there has been intense interest in the design of bacteriotherapies or ``bugs as drugs,'' which are communities of bacteria administered to patients for specific therapeutic applications. Central to the design of such therapeutics is an understanding of the causal microbial interaction network and the population dynamics of the organisms. In this work we present a Bayesian nonparametric model and associated efficient inference algorithm that addresses the key conceptual and practical challenges of learning microbial dynamics from time series microbe abundance data. These challenges include high-dimensional (300+ strains of bacteria in the gut) but temporally sparse and non-uniformly sampled data; high measurement noise; and, nonlinear and physically non-negative dynamics. Our contributions include a new type of dynamical systems model for microbial dynamics based on what we term interaction modules, or learned clusters of latent variables with redundant interaction structure (reducing the expected number of interaction coefficients from $O(n^2)$ to $O((\log n)^2)$); a fully Bayesian formulation of the stochastic dynamical systems model that propagates measurement and latent state uncertainty throughout the model; and introduction of a temporally varying auxiliary variable technique to enable efficient inference by relaxing the hard non-negativity constraint on states. We apply our method to simulated and real data, and demonstrate the utility of our technique for system identification from limited data, and for gaining new biological insights into bacteriotherapy design.
\end{abstract}

\section{Introduction}
\begin{figure}[t!]
\centering
\includegraphics[width=.47\textwidth]{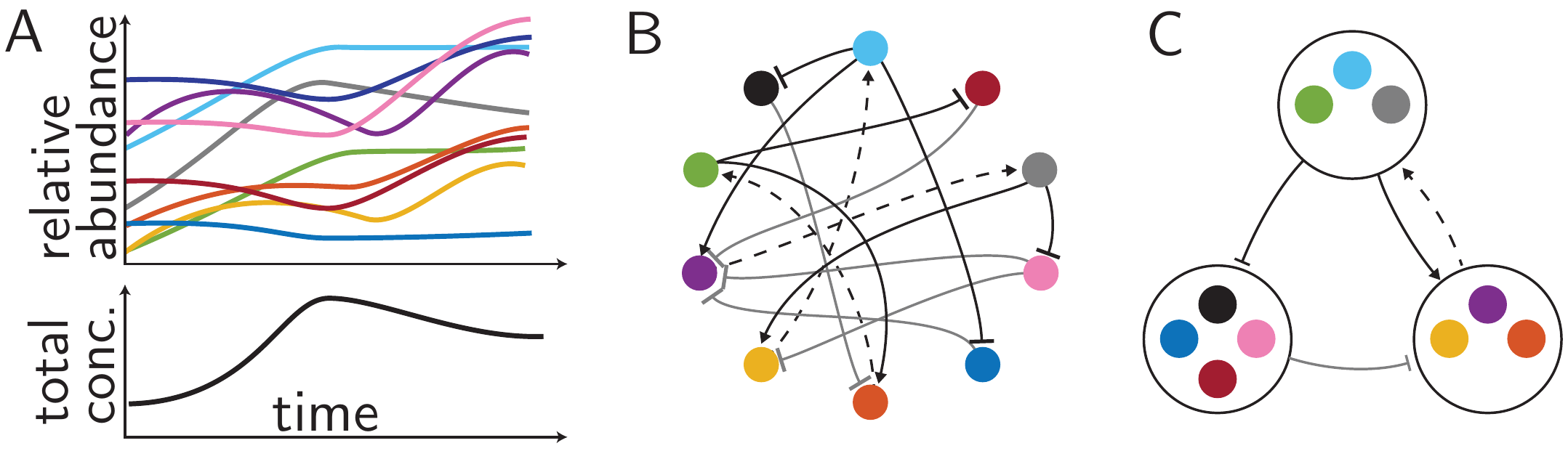}\vspace{0pt}
\caption{Schematic illustrating task of dynamical systems inference from microbiome time series data: \textbf{(A)} Input is time series of relative abundances of microbial species and time series of total microbial concentrations \textbf{(B)} Pairwise microbe-microbe interaction network reflecting non-zero interaction coefficients in underlying dynamical systems model. \textbf{(C)} Microbe-microbe interaction network with interaction module structure.}\label{fig:schematic}
\end{figure}

The human microbiome constitutes all the microorganisms that live in and on our bodies \cite{:2012aa}.
There is strong evidence that the microbiome  plays an important role in a variety of human diseases, including:
infections, arthritis, food allergy, cancer, inflammatory bowel disease, neurological diseases, and obesity/diabetes \cite{hall2017human,Youngster01062014,Stefka09092014,schwabe2013microbiome,kostic2015dynamics,wlodarska2015integrative}. Given the microbiome's profound role, there is now a concerted effort to design bacteriotherapies, which are cocktails of multiple bacteria working in concert to achieve specific therapeutic effects. Multiple strains are often needed in bacteriotherpies both because multiple host pathways must be targeted, and because additional bacteria may provide stability or robustness to the community as a whole. An important step toward designing bacteriotherapies is mapping out microbial interactions and predicting population dynamics of this ecosystem. One approach toward this goal, and arguably the most popular, is to learn dynamical systems models from time series measurements of microbiome abundance data. That is, one takes as input time series of microbiome abundances as depicted in Figure \ref{fig:schematic}A and infers a dynamical systems model of microbial interactions as in Figure \ref{fig:schematic}B. These data typically consist of two separate measurements: (1) high-throughput next generation sequencing counts of a marker gene (16S rRNA) mapped back to different microbial species or other taxonomic units (often 300+), to determine relative abundances of each unit, and (2) quantitative PCR (qPCR) measurements to determine the total concentration of bacteria in the ecosystem.

Inferring dynamical systems models from microbiome time series data presents several challenges. The biggest challenge arises from the fact that the data is high-dimensional, yet temporally sparse and non-uniformly sampled. With 300 or more bacterial species in the gut, the resulting differential equation models can have more than 90,000 possible interaction parameters. However, unlike other biomedical domains where almost continuous temporal sampling is feasible (e.g., electrical recordings of cardiac activity), this is not currently possible for the gut microbiome. Instead, we must rely on fecal samples (or even more invasive processes, such as colonoscopy), which means that we are quite limited in terms of the frequency and total number of samples. Further, the techniques used to obtain estimates of microbial abundance are noisy, and with multiple technologies being combined (i.e., next generation sequencing and qPCR), the resulting measurement error models are relatively complex. Finally, the microbiome exhibits nonlinear and physically nonnegative dynamics, which introduce additional inference issues. 

\subsection{Prior work}

We now briefly review previous work in inferring dynamical systems from microbiome time-series data. The authors of \cite{stein:2013} model microbial dynamics using continuous time deterministic generalized Lotka-Volterra (gLV) equations, transform to a discrete time linear model via a log transform to enable efficient inference, and then use L2 penalized linear regression to infer model parameters. The transformation performed in \cite{stein:2013} is common in the ecological literature, and provides a point of comparison to our model, so we present it in detail now. Deterministic gLV dynamics can be written compactly as the Ordinary Differential Equation (ODE) $\dot x(t) = x(t) \odot(r+Ax(t))$, where $\odot$ is the element wise product for vectors, $r$ is a vector of growth rates and $A$ is a matrix of interaction coefficients. Using $\oslash$ for element wise division, the following representation of the ODE also holds: $\dot x(t)\oslash x(t) = r+Ax(t)$. The left hand side of the equivalent ODE can then be integrated resulting in the following identity: $\int_{t_1}^{t_2} \dot x(t)\oslash x(t) \, \mathrm d t = \log(x(t_2))- \log(x(t_1))$. This property of the logarithm can then be used to approximate the continuous time nonlinear ODE as a discrete time linear dynamical system. There are a variety of both theoretical and practical issues with using this approximation. For instance, the transformation does not readily apply for stochastic dynamics. Additionally, the transform essentially assumes normally distributed error, which is inherently false, since data typically consist of sequences of counts. Further, we often encounter measurements of zero for microbial abundance, i.e., below the limit of detection, which would lead to taking the $\log$ of zero or adding an artificial small number.

Other work on inferring dynamical systems models from microbiome data includes \cite{10.1371/journal.pone.0102451}, which takes a similar approach to \cite{stein:2013},  but instead of L2 penalized regression, use a sparse linear regression with bootstrap aggregation approach. No regularization is performed and sparsity is introduced into the model by adding and removing interaction coefficients one at a time with step-wise regression. Several inference techniques are presented in \cite{Bucci2016}, two being extensions of the model proposed in \cite{stein:2013} and two being new Bayesian models. The Bayesian models in \cite{Bucci2016} are based on ODE gradient matching, in which Bayesian spline smoothing is first performed to filter the experimental measurements, and then a Bayesian adaptive lasso or Bayesian variable selection method is used to infer model parameters. These methods do incorporate non-normally distributed measurement error models, but errors are not propagated throughout the model, i.e., smoothing and filtering are separate steps. Finally, in \cite{alshawaqfeh2017inferring} an Extended Kalman Filter (EKF) is applied to a stochastic gLV model, which incorporates filtering directly, unlike the aforementioned references; however, noise is assumed to be normally distributed.

Beyond microbiome specific dynamical systems inference approaches, there is an extensive body of work on Bayesian inference of nonlinear dynamical systems, which remains an active area of research \cite{Ionides05122006,carlin1992monte,aguilar1998bayesian,geweke2001bayesian}. An interesting line of recent work leverages Gaussian Processes (GP) as a means for efficient filtering for both ordinary differential equations and partial differential equations \cite{chkrebtii2016bayesian}. One of the catalysts for this line of work came from \cite{calderhead2009accelerating}, in which a GP is used to infer the latent state variables, which in turn are used to infer parameters of an ODE. Extending that work, \cite{dondelinger2013ode} apply a gradient matching approach (marginalizing over state derivatives) and perform joint inference on the ODE parameters and latent state variables. However, several subsequent papers pointed out identifiability and efficiency issues with these approaches
\cite{barber2014gaussian,macdonald2015controversy}. More recently, \cite{NIPS2017_7066,NIPS2017_7274} presented a variational inference approach that addresses some of these issues. While we do not explore GPs in this work, they are an interesting and promising direction within the broader domain of Bayesian inference for nonlinear dynamical systems. Dynamic Bayesian Networks (DBN)  also represent a broad class of state-space models leveraged for inference of dynamical systems given time series data \cite{murphy2002dynamic}.  Our model differs from a standard DBN, in that it learns the conditional independence structure in a latent temporal space, and clusters the nodes in the graph nonparametrically.

Also related to our work are models that learn clustered representations of interacting systems, both for purposes of enhancing interpretability and for increasing efficiency of inference. Related approaches include Stochastic Block Models (SBM), in particular \cite{kemp2006learning}, which model redundant interaction structure as probabilistic linkages between individual actors that are influenced by the blocks/groups that the actors belong to. SBMs  typically directly model observed, non-temporal data, whereas our approach models latent temporal signals; further, our approach enforces identical interaction structure on variables in the same cluster, whereas SBMs assume a probabilistic interaction structure. Dependent groups/clusters have also been explored in the context of Topic Models (e.g., \cite{mimno2007mixtures}). There is also an extensive literature on Dependent Dirichlet Processes \cite{maceachern2000dependent}, which can be used to capture complex interactions between clusters, and also simpler structures (e.g., hierarchies as in \cite{teh2006hierarchical}). 

\subsection{Contributions}
In this work we present a Bayesian nonparametric model and associated efficient inference algorithm that addresses the key conceptual and practical challenges of learning microbial dynamics from time series microbe abundance data. Our main contributions are:
\begin{itemize}
	\item A new type of dynamical systems model for microbial dynamics based on what we term \textit{interaction modules}, or probabilistic clusters of latent variables with redundant interaction structure. The aggregated concentrations of microbes in a module act as consolidated inputs to other modules, with structural learning of the network of interactions among modules.
	\item A fully Bayesian formulation of the stochastic dynamical systems model that propagates measurement and latent state uncertainty throughout the model. This integrated approach improves on the previous work described for microbiome dynamics (which assumed deterministic dynamics and separated learning of latent states and ODE parameters).
	\item Introduction of a temporally varying auxiliary variable technique to enable efficient inference by relaxing the hard non-negativity constraint on states. Introduction of the auxiliary variable not only allows for efficient inference with respect to filtering the latent state, it also allows for collapsed Gibbs sampling for module assignments and for the structural network learning component. 
\end{itemize}

The remainder of this paper is organized as follows. In Section 2 we present the complete model. Section 3 describes our inference algorithm. Section 4 contains experimental validation on simulated and real data. Section 5 contains our concluding remarks. Before moving on, a quick comment regarding notation: random variable are written in bold as $\bm \alpha, \bm \beta,  \bm \gamma, \p a, \p b, \p c$ with regular parameters denoted as $\alpha, \beta, \gamma, a, b, c$.

\section{Model}
\subsection{Model of dynamics}
Our model of dynamics is based on a stochastic version of the gLV equations, widely used in ecological system modeling:
\ben
\mathrm d \p x_{t,i} = \p x_{t,i} \bigl(a_{i,1} + a_{i,2} \p x_{t,i} +  \textstyle\sum\nolimits_{j \neq i} b_{ij} \p x_{t,j} \bigr) \mathrm dt + \mathrm d \p w_{t,i} 
\een%
$i \in \{1,2,\ldots,n\}$ where $\p x_{t,i}\in \Re_{\geq 0}$ is the abundance of microbial species $i$ at time $t\in \Re$,  $a_{i,1}\in\Re$ is  the growth rate of microbial species $i$ and $a_{i,2}$ is the ``self interaction term'' and together $a_{i,1}$ and $a_{i,2}$ determine the carrying capacity of the environment when species $i$ is not interacting with any other species. The  coefficients $b_{ij}$ when $i\neq j$ are then the microbial interaction terms. The term $\p w_{t,i} \in \Re$ represents a stochastic disturbance. Note that, while not shown explicitly, the disturbance must be conditioned on the state to prevent negative state values. Overloading the first subscript in $\p x$, a discrete-time approximation to the gLV dynamics above is:%
\begin{multline}
\label{linearglv}
\p x_{(k+1),i} - \p x_{k,i} \approx  \p x_{k,i} \bigl(a_{i,1} + a_{i,2} \p x_{k,i} + \textstyle\sum\nolimits_{j \neq i} b_{ij} \p x_{k,j} \bigr) \Delta_k \\+ \sqrt{\Delta_{k}} (\p w_{{k+1},i} - \p w_{k,i}) 
\end{multline}%
where $k\in \mathbb N_{>0}$ indexes time as $t_k$ and $\Delta_k \triangleq t_{k+1} - t_k$. 

The accuracy of this approximation will depend on a sufficiently dense discretization relative to time-scales of the dynamics of interest. Higher order integration methods are possible for Stochastic Differential Equations (SDE), but quickly become very complicated without straightforward gains in accuracy seen with ODEs. Our experience has been that Euler methods behave well for the gLV model in real microbial ecosystems, which are inherently stable. However, Euler integration may be sub-optimal for strongly perturbed systems (e.g., antibiotics). We note that Euler integration is indeed an advance over the state-of-the-art, which uses gradient-matching methods that don't perform any integration. An interesting area for future work would be to leverage Bayesian Probabilistic Numerical Methods \cite{cockayne2017bayesian} to incorporate step-size adaptation directly into our model.

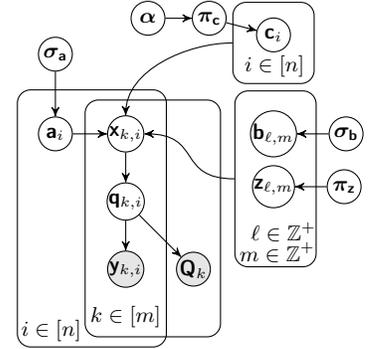
\begin{figure*}[th!] \centering
\begin{minipage}[T]{0.71\textwidth}
\begin{tcolorbox}[width=3.8in,
colframe=black,
colback=white,boxrule=0.5pt,
                  boxsep=0pt,
                  left=0pt,
                  right=0pt,
                  top=2pt]
                  \small
\text{ Dirichlet Process} \hspace{0.99in} \text{Edge Selection}%
\setlength{\abovedisplayskip}{2pt}%
\setlength{\belowdisplayskip}{2pt}%
\begin{align*}
\bm \pi_\p c\mid \bm \alpha &\sim \mathtt{Stick}(\bm \alpha) &  \p z_{\p c_i,\p c_j} \mid \bm\pi_\p z&\sim \mathtt{Bernouli}(\bm \pi_\p z)  \\
\p c_i \mid \bm \pi_\p c&\sim \mathtt{Multinomial}(\bm\pi_\p c) & \text{ Self Interactions} \\
 \p b_{\p c_i,\p c_j} \mid \bm \sigma_\p b&\sim \mathtt{Normal}(0,\bm \sigma_\p b^2) &  \p a_{i,1},  \p a_{i,2} \mid \bm\sigma_\s a  & \sim \mathtt{Normal}(0,\bm \sigma_\p a^2)
 \end{align*}
{ Dynamics}
\begin{multline*}
\p x_{k+1,i} \mid \s x_k,\s a_i, \s b, \s z, \s c, \bm \sigma_\p w  \sim 
\\ \hspace{.4in}\mathtt{Normal}\Bigl(\p x_{k,i} +\p x_{k,i} \Bigl(\p a_{i,1} + \p a_{i,2} \p x_{k,i} + \textstyle\sum\limits_{\mathclap{\p c_j\neq \p c_i}}  \p b_{\p c_i,\p c_j}  \p z_{\p c_i,\p c_j} \p x_{k,j} \Bigr), \Delta_k\bm\sigma_\p w^2\Bigr)
\end{multline*}
{ Constraint and Measurement Model}
\begin{align*}
\p q_{k,i} \mid \s x_{k,i} &\sim \mathtt{Normal}(\p x_{k,i},\sigma_\p q^2)  \\
\p y_{k,i}\mid  \p q_{k,i} & \sim \mathtt{NegBin} (\phi(\p q_{k}),\epsilon(\p q_{k})), \quad \phi,\ \epsilon \text{ defined in \eqref{eq:phi}, \eqref{eq:eps}}\\
\p Q_{k}\mid  \p q_{k,i} & \sim  \mathtt{Normal}\left (\textstyle{\sum_ i} \s q_{k,i}, \sigma_{\s Q_k}^2 \right ) 
\end{align*}
\end{tcolorbox}
\end{minipage}\hspace{1pt}
\begin{minipage}[T]{0.25\textwidth}
\begin{tikzpicture}[>=stealth',scale=0.9,every node/.style={scale=0.9},auto,node distance=1cm,cir/.style={circle,draw},
 minimum size=.5cm,inner sep=0pt,baseline=(current  bounding  box.center),red node/.style={circle,draw,color=gray,fill},squ/.style={draw},font=\footnotesize]

  \node[cir] (x1)  {$\p x_{k,i} $};
  \node[cir] (q1) [below of=x1] {$\p q_{k,i}$};
  \node[cir] (y1)[below of=q1,fill=black!10]   {$\p y_{k,i}$};
    \node[cir] (Q1)[right of=y1,fill=black!10]   {$\p Q_{k}$};

   	\node[cir,minimum size=.5cm] (b) [right=40pt of x1] {$\p b_{\ell,m}$};
       	\node[cir,minimum size=.5cm] (sib) [right=12pt of b] {$\bm\sigma_\p b$};
        \node[cir,minimum size=.5cm] (z) [below=3pt of b] {$\p z_{\ell,m}$};
      	\node[cir,minimum size=.5cm] (pia) [right=12pt of z] {$\bm \pi_\p z$};
	\node[cir,minimum size=.5cm] (a) [left=13pt of x1] {$\p a_i$};

	\draw[rounded corners=5pt] (-0.6,-3) rectangle (1.4,0.5);
	\node at (0,-2.7)  {$k\in[m]$};
	\draw[rounded corners=5pt] (-1.6,-3.15) rectangle (.6,0.65);
	\node at (-1.1,-2.9)  {$i\in[n]$};
      	\node[rectangle,draw,minimum width=1.2cm,minimum height=2.6cm,rounded corners=5pt]  (r3) [above right=-30pt and 36pt of q1] {};
        \node[cir,minimum size=.5cm] (c) [above=22pt of b] {$\p c_i$};
        	\node[cir,minimum size=.5cm] (pic) [above left=-3pt and 15pt of c] {$\bm \pi_\p c$};
	\node[cir,minimum size=.5cm] (alpha) [left=10pt of pic] {$\bm \alpha$};
        \node[cir,minimum size=.5cm] (nu) [above=17pt of a] {$\bm \sigma_\p a$};
        \node[rectangle,draw,minimum width=1.2cm,minimum height=1.2cm,rounded corners=5pt]  (r4) [above=10pt of b] {};
	\node at (2.33,-1.45)  {$\ell\in\mathbb Z^+$};
        \node at (2.24,-1.75)  {$m\in\mathbb Z^+$};
        \node at (2.2,1.01)  {$i\in[n]$};
      
      	\path[->]
	(alpha) edge (pic)
  	(x1) edge (q1)
	(q1) edge (Q1)
  	(q1) edge (y1)
     	(r4) edge[out=180, in=90,looseness=.9] (x1)
      	(nu) edge (a)
    	(pia) edge (z)
 	(sib) edge (b)
  	(pic) edge (c)
	(a) edge (x1)
	(r3) edge[out=180, in=0,looseness=1.5] (x1);
\end{tikzpicture}
\end{minipage}\vspace{-2pt}
\caption{Mathematical description of the model and the graphical model. Higher level priors are not depicted in the model.}
\label{fig:block}
\end{figure*}

\subsection{Interaction modules}
 We incorporate a Dirichlet Process (DP)-based clustering technique \cite{neal2000markov,rasmussen2000infinite} to learn redundant interaction structures among bacterial species, which we term interaction modules. In the context of our dynamical systems model, this means that only interaction coefficients between modules need to be learned, rather than interactions between each pair of microbes. Without modules, the number of possible interaction coefficients scales as $O(n^2)$, where $n$ is the number of microbial species. Since we are using DPs, where the expected number of clusters is $O(\log n)$ \cite{antoniak1974mixtures}, the expected number of interaction coefficients is $O((\log n)^{2})$. For purposes of interpretability, we specifically assume no interactions within each module, corresponding to the biologically important scenario of redundant functionality among sets of microbes. An example of interaction module structure is visualized in Figure \ref{fig:schematic}C: while Figures \ref{fig:schematic}B and \ref{fig:schematic}C both contain 10 microbes, there are only 6 interactions to learn in \ref{fig:schematic}C (between modules), versus 90 microbe-microbe interactions in \ref{fig:schematic}B without the module structure. 

Figure \ref{fig:block} depicts our interaction module model as a generative model. Starting with the Dirichlet Process, ${\p c_i\in \mathbb Z^+}$ represents the cluster assignment for bacterial species $i$. If species $i$ and species $j$ are in different clusters, and thus $\p c_i \neq \p c_j$, then ${\p b_{\p c_i,\p c_j}\in\Re}$ is the coefficient representing the (interaction) effect that the module containing species $j$ has on species $i$. If species $\ell$, different from species $i$, is in the same cluster as species  $j$,  then $\p b_{\p c_i,\p c_j}=\p b_{\p c_i,\p c_{\ell}}$ by definition (i.e., species in the same cluster share interaction coefficients). Note that no interactions are assumed to occur within a cluster, as discussed. 

For each element in $\p b$ there is a corresponding element in $\p z$, which is an indicator variable (0 or 1) that chooses whether an interaction exists between two modules. Thus, our model automatically adapts the interaction network by structurally adding or removing edges (analogous to approaches for standard Bayesian Networks e.g., \cite{george1993variable,Heckerman2008}), which we refer to as Edge Selection (ES). This approach allows us to easily compute Bayes factors \cite{kass1995bayes}, enabling principled determination of the evidence for or against each interaction occurring.

The terms $\p a_{i,1}$ and $\p a_{i,2}$ correspond to the growth rate and self interaction term for species $i$, respectively. Note that these variables are not part of our clustering scheme and do not have indicator variables associated with them.

\subsection{Modeling non-negative dynamics}
{We now discuss one of our technical contributions, which is to relax the strict non-negativity assumption on $\p x$ in Equation \eqref{linearglv} and thereby enable efficient inference while maintaining (approximate) physically realistic non-negative dynamics. To accomplish this, we introduce an auxiliary trajectory variable $\p q$ such that $\p q_{k,i} \sim \mathtt{Uniform}[0,L)$, with $L>0$ and much larger than any of the measured values. Microbial abundance data $\p y$ are assumed to be generated from $\p q$ through some model of measurement noise $\p y \mid \p q$ (discussed below).
	
We couple the latent trajectory $\p x$ to the auxiliary variable $\p q$ through a conditional distribution $\p q \mid \p x$, which we assume to be Gaussian with small variance. This effectively introduces a momentum term into the model of dynamics \eqref{linearglv} (proportional to the difference between $\p x$ and $\p q$), which softly constrains $\p x$ to be in the range $[0,L)$. This renders the posterior distributions for $\p x$ and gLV parameters $\s a,\s b$ Gaussians rather than their being truncated Gaussians if strict non-negativity were imposed. Our technique has connections to several approaches that break or relax dependencies in a model to improve inference efficiency, such as Variational Inference \cite{doi:10.1080/01621459.2017.1285773} and distributed/parallel Bayesian inference approaches \cite{angelino2016patterns}. 

\begin{figure}[tb!]\center
\begin{tikzpicture}[>=stealth',auto,node distance=1.2cm,cir/.style={circle,draw},
 minimum size=.5cm,inner sep=0pt,baseline=(current  bounding  box.center),red node/.style={circle,draw,color=gray,fill},squ/.style={draw},font=\small,scale=0.75, every node/.style={transform shape}]

  \node[cir] (z31) [fill=ml1!40] {$\s x_1$};
  \node[cir] (z32) [right of=z31,fill=ml5!40] {$\s x_2$};
  \node[cir] (z33) [right of=z32,fill=ml2!40] {$\s x_3$};
  \node(z3d) [right of=z33,node distance=1.0cm] {$\cdots$};
  \node[cir] (z3n) [right of=z3d,node distance=1.0cm] {$\s x_n$};
  \node[cir,minimum size=.5cm,fill=ml1!40] (w3) [above left=0.2cm and .5cm of z31] {$\s a$};
  \path[->]    
    (z31) edge (z32)
    (z32) edge (z33)
    (z33) edge (z3d)
    (z3d) edge (z3n)
    (w3) edge[out=0, in=135,looseness=.7] (z32)
    (w3) edge[out=0, in=135,looseness=.5] (z33)
    (w3) edge[out=0, in=135,looseness=.3] (z3n);
  \node[cir] (y1)[below of=z31,node distance=1.2cm]   {\color{black}$\s q_1$};
  \node[cir] (y2) [right of=y1,fill=ml1!40] {\color{black}$\s q_2$};
  \node[cir] (y3) [right of=y2] {\color{black}$\s q_3$};
  \node(yd) [right of=y3,node distance=1.0cm] {$\cdots$};
  \node[cir] (yn) [right of=yd,node distance=1.0cm] {\color{black}$\s q_n$};

  \path[->]
  (z31) edge (y1)
    (z32) edge (y2)
     (z33) edge (y3)
      (z3n) edge (yn);
      
        \node[cir,fill=black!10] (y11)[below of=y1,node distance=1.2cm]   {$\s y_1$};
  \node[cir,fill=black!10] (y21) [right of=y11] {$\s y_2$};
  \node[cir,fill=black!10] (y31) [right of=y21] {$\s y_3$};
  \node(yd1) [right of=y31,node distance=1.0cm] {$\cdots$};
  \node[cir,fill=black!10] (yn1) [right of=yd1,node distance=1.0cm] {$\s y_n$};
    \path[->]
 (y1) edge (y11)
    (y2) edge (y21)
     (y3) edge (y31)
      (yn) edge (yn1);
    \end{tikzpicture}\vspace{-4pt}
    \caption{Our model unrolled-in-time to explicitly show temporal dependencies. Color coding (blue, green, orange) used to visualize our proposal distribution when filtering latent state $\s x$, see \S 3. }\label{fig:time}
    \end{figure}
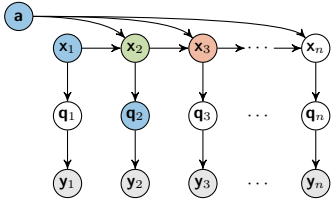

Our approach can also be thought of as a product of experts: one expert is a uniform distribution confining $\p q$ to the positive orthant, and the other is a normal distribution enforcing closeness to the actual trajectory $\p x$. With either interpretation, $\p q$ acts as a ``restoring force'' that pulls the posterior of $\p x$ toward the positive orthant. With the introduction of $\s q$, the posterior $\s a \mid \s x$ is now simply a multivariate Gaussian. Practically, this makes efficient inference feasible, since sampling from the posterior is now easy and we can also perform closed-form marginalizations. Further, the measurement model is decoupled from the dynamics, allowing for efficient inference with flexible measurement noise models, such as negative binomial distributions for modeling sequencing counts \cite{paulson2013differential,love2014moderated}. This is explored in detail in the subsequent subsection. In the Appendix, we provide a detailed analysis of the issues that ensue with a naive model that directly enforces non-negativity through the dynamics.

\subsection{Measurement Model}
Our measurement model handles two experimental technologies, sequencing counts of a marker gene (16S rRNA) mapped back to different microbial species or other taxonomic units, and qPCR measurements to determine total microbial concentration in the sample.  The variable $\s y_{k,i}$ denotes the number of counts (sequencing reads) associated with bacterial species $i$ at time $k$ and $\s Q_k$ is the total bacterial concentration at time $k$. Our complete sensor model combining the two measurements is illustrated in Figure \ref{fig:block}. The counts measurements $\s y_{k,i}$ are sampled from a Negative Binomial distribution with mean and dispersion parameters defined as:
\begin{align}
\s y_{k,i} \mid \s q_k & \sim \mathtt{NegBin}(\phi(\s q_k,r_k) ,\epsilon(\s q_k, a_0,a_1) ) \nonumber \\
\phi(q_k, r_k) &=  r_k \frac{q_{k,i}}{\sum_i q_{k,i}}  \label{eq:phi}\\
\epsilon(q_k, a_0,a_1)&= \frac{a_0}{q_{k,i}/{\sum_i q_{k,i}} }+a_1   \label{eq:eps}
\end{align}
where $r_k$ is the total number of sequencing reads for the sample at time $k$ (often referred to as the read depth of the sample). The form of this model follows that of \cite{Bucci2016,love2014moderated}; see these references for detailed discussions on the validity of, and the empirical evidence for, using this error model for next generation sequencing counts data.

The Negative Binomial dispersion scaling parameters $a_0,a_1$ are pre-trained on raw reads, and are not learned jointly with the rest of the model. Similarly, measurement variance, $\sigma^2_{\s Q_k}$ is estimated directly from technical replicates for each measurement. For completeness, we also give our specific parameterization of the Negative Binomial Probability Density Function (PDF):
\begin{align*}
 \mathtt{NegBin}(y; \phi ,\epsilon ) =& \frac{\Gamma(r+y)}{y!\,\Gamma(r)} \left( \frac{\phi}{r+\phi} \right)^y \left( \frac{r}{r+ \phi}\right)^{r} \\
r =& \frac{1}{\epsilon} 
\end{align*}
With this parameterization of the Negative Binomial distribution, the mean is $\phi$ and the variance is  $\phi + \epsilon \phi^2$.

\subsection{Additional priors not specified in Figure \ref{fig:block}}\label{sec:prior}
To complete the model description, we describe higher-level priors not shown in Figure \ref{fig:block}. For the three variance random variables $(\bm \sigma^2_\s a,\bm \sigma^2_\s b,\bm \sigma^2_\s w)$ Inverse-Chi-squared priors are used. The concentration parameter $\bm \alpha$ for the DP is given a Gamma prior. Hyperparameters were set using a technique similar to \cite{Bucci2016}, where means of distributions were empirically calibrated based on the data and variances were set to large values to produce diffuse priors.

\begin{figure*}[t!]
\centering
\includegraphics[width=.8\textwidth]{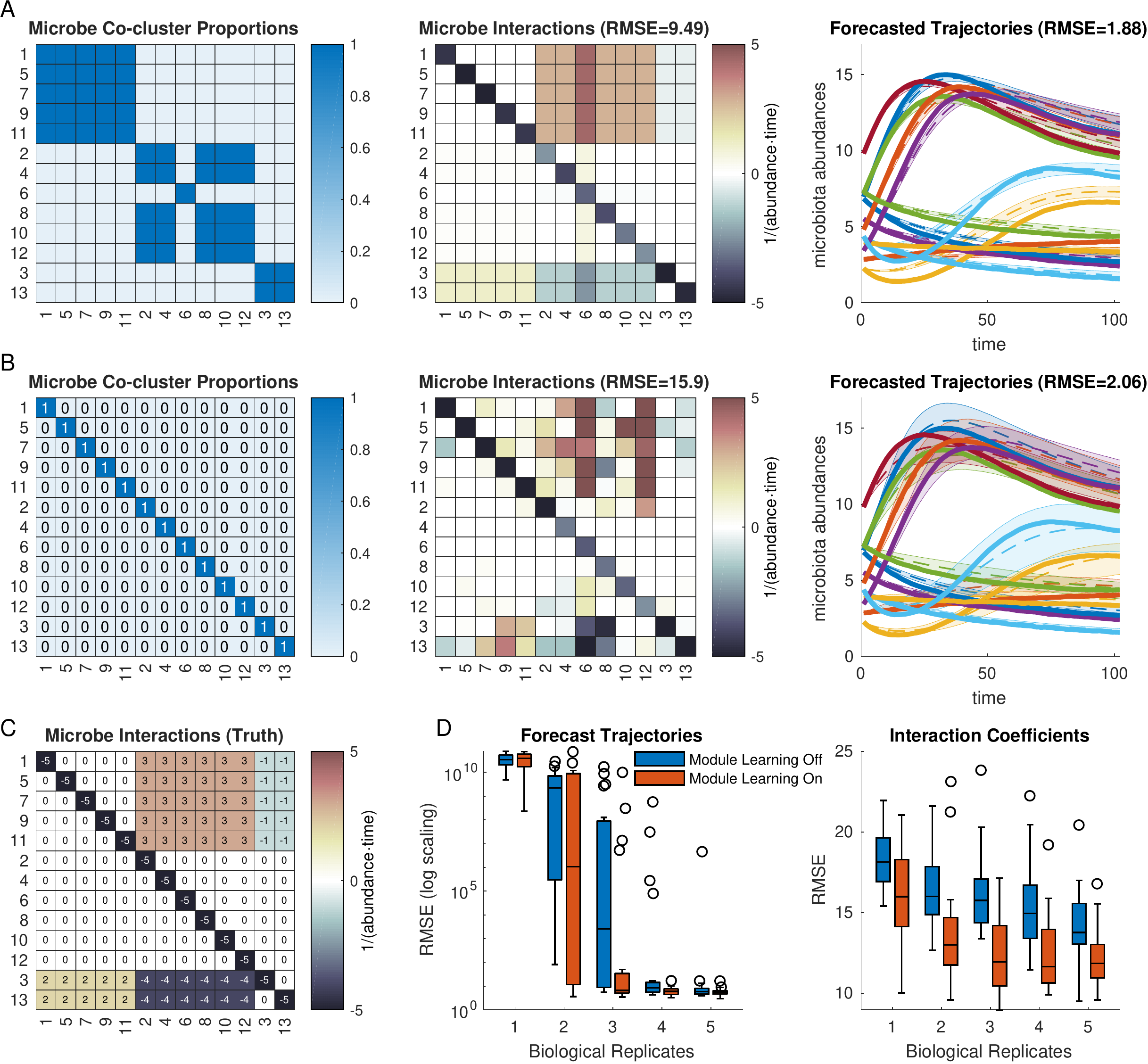}\vspace{-8pt}
\caption{Results on simulated data with (and without) interaction module learning.  Module learning greatly improves accuracy in terms of identifying ground truth interaction coefficients.
 With enough biological replicates, both methods have similar performance in terms of forecasting microbial abundance trajectories. ({\bf A}) Inference with interaction module learning enabled. ({\bf left}) Co-cluster proportions illustrating the probability that two microbes appear in the same module. ({\bf middle}) Expected values for interaction coefficients.  ({\bf right}) Forward simulated dynamics from initial conditions not in the training set. Ground truth microbe abundance trajectory shown as solid line. 95\% intervals shown as shaded regions with the expected trajectory shown as a dashed line. ({\bf B}) Inference without interaction module learning enabled. ({\bf C}) Ground truth interaction matrix, which also illustrates the underlying simplified interaction structure of the graph in \eqref{graph:synth1}. ({\bf D}) Forecasting microbial abundance trajectories and interaction coefficient inference performed 20 times for a range of numbers of biological replicates $\{1,2,\ldots 5\}$. Shaded boxes denote 25th and 75th percentile, the solid line is the median, whiskers constructed from 1.5 times the interquartile region, and outliers shown as circles. Large RMSE in forecasting arises from the fact that without sufficiently rich data the model learns coefficients that do not result in stable dynamics.}\label{fig:synth}
\end{figure*}

\section{Inference}

We briefly describe our Markov Chain Monte Carlo inference algorithm, which leverages efficient collapsed Gibbs sampling steps.  As described in Section \ref{sec:prior}, we use conjugate priors on many variables (e.g., the variance terms $(\bm \sigma^2_\s a,\bm \sigma^2_\s b,\bm \sigma^2_\s w)$), which allows straight-forward Gibbs sampling. The module assignments, $\p c$, are also updated by a standard Gibbs sampling approach for Dirichlet Processes~\cite{neal2000markov}. For the concentration parameter $\bm \alpha$, which has a Gamma prior on $\bm \alpha$, we use the sampling method described by \cite{escobar1995bayesian}.

Our auxiliary trajectory variables $\p q$ allow us to marginalize out in closed form the interaction coefficients $\p b$, and thus perform collapsed Gibbs sampling, both during sampling assignments of species to modules and when structurally learning the network of interactions between modules. Collapsed Gibbs steps have been shown to improve mixing substantially for DP inference~\cite{neal2000markov}.

Sampling of the auxiliary variables $\p q$ and latent trajectories $\p x$ require Metropolis-Hastings (MH) steps. Briefly, for $\p q$, the MH proposal is based on a Generalized-Linear Model approximation. For $\p x$, we use a one time-step ahead proposal similar to that described in \cite{geweke2001bayesian}. Our proposal uses the previous time point latent abundance, the gLV coefficients, and the auxiliary trajectory (which is directly coupled to the observations) to propose the next time point abundance giving the proposal the form $p_{\p x_{k+1} \mid \p x_k, \p q , \bm \Omega}$, where $\bm \Omega=\s a_i, \s b, \s z, \s c, \bm \sigma_\p w $.  Thus, our proposal is essentially the forward pass of a Kalman filter (which we color coded in Figure \ref{fig:time}).  Our proposal uses the information from the blue nodes, to propose for the green node. The future state information (orange node) is not used for the proposal, for efficiency of computation (i.e., we exploit conjugacy for the forward pass). The future state information comes into the target distribution, so we sample from the true posterior. Note that this is different from a standard Extended Kalman Filtering approach, which linearizes around estimated mean and covariance and can deviate substantially from the true posterior.

\section{Results}

In this section we present results applying our model to both simulated and real microbiome data. Our goal with simulated data is to illustrate the utility of our model (and specifically Module Learning) when  inferring microbial dynamics from time series data with limited biological replicates and temporal resolution, which is the reality for \emph{in vivo} microbiome experiments. Figures \ref{fig:synth}A-\ref{fig:synth}C depict our results, comparing inference both with and without interaction module learning. Simulated data was constructed to mimic state-of-the-art experiments for developing and testing bacteriotherapies \cite{Bucci2016}. In these experiments, germ-free mice (animals raised in self-contained bacteria-free environments) were inoculated with defined collections of 13 bacterial species and serial fecal samples were collected to analyze dynamics of microbial colonization over time. Due to costs and logistic constraints, such experiments use relatively small numbers of biological replicates ($\approx 5$ mice) and limited temporal sampling (e.g., 10-30 time-points per mouse). To simulate these experiments, we generated data with 5 biological replicates (5 different time series simulated from the same dynamics, but with different initial conditions), 11 time-points per replicate, and assumed gLV dynamics with the following module and interaction structure: 
\begin{equation}
\begin{tikzpicture}[>=stealth',scale=1,every node/.style={scale=1},auto,node distance=1cm,cir/.style={circle,draw},
 minimum size=28pt,inner sep=0pt,baseline=(current  bounding  box.center),red node/.style={circle,draw,color=gray,fill},squ/.style={draw},font=\tiny]

  \node[cir] (a) at (90:1.2)[text width=25pt] {$\, \  1,5,7$\\$\, \ \  9,11$};
  \node[cir] (b) at (210:1.2) [text width=25pt] {$2,4,6,8$\\ \ \ $10,12$};
  \node[cir] (c) at (330:1.2) [text width=25pt]   {$\  \  \, 3,13$};
   	      
      	\path[->]
	(a) edge[out=320, in=100] node[right,xshift=-3mm,yshift=1mm] {$2$} (c) 
  	(b) edge node[below,yshift=2mm] {$-4$} (c)
  	(b) edge node[left,xshift=2mm,yshift=1mm] {$3$} (a)
     	(c) edge[out=135, in=285]  node[left,xshift=2mm,yshift=-1mm] {$-1$}(a);
\end{tikzpicture}\label{graph:synth1}\vspace{0pt}
\end{equation}
where the numbers inside the nodes represent bacterial species in the same module and the edge weights are the module interaction coefficients $\s b_{\s c_i, \s c_j}$ in our model in Figure \ref{fig:block}. Note that this graph in \eqref{graph:synth1} is just another representation of the weighted adjacency matrix in Figure \ref{fig:synth}C.

With module learning (Figure \ref{fig:synth}A), our algorithm recovers the module structure as expected, almost completely correctly, and also recovers the interaction coefficients well. While the algorithm incorrectly places species 6 in its own cluster, it properly learns that no other species contribute to the dynamics of species 6 (i.e. elements in  the row associated with species 6, other than the self interaction term, are zero). Our algorithm also forecasts trajectories of microbial abundances quite accurately. Without module learning enabled (Figure \ref{fig:synth}B), the  algorithm still forecasts trajectories fairly accurately (although slightly worse than with module learning), but does much worse in inferring the interaction coefficients, and indeed the actual structure of the dynamical system is not at all evident. The ability to forecast trajectories relatively accurately, but not recover the underlying structure of the system well, highlights the issues with identifiability of nonlinear dynamical systems models from limited data: without additional structural constraints in the model, it is too easy to overfit, because many different settings of ODE parameters can result in exactly the same trajectories. 

To investigate this issue further, we performed additional simulations using the same setup with varying numbers of biological replicates (Figure \ref{fig:synth}D). Results using 20 initial conditions were run and aggregate statistics are presented. For forecasting trajectories, module learning clearly helps, although performance is relatively good without module learning with 4 or more biological replicates. However, as can be seen, for identification of the actual ODE parameters, module learning has a much larger advantage.

It is worth noting that module learning also resulted in significant improvements in wall-clock runtime, by a factor of about 10. We did not test this empirical observation extensively, but it is consistent with theory, in that the additional time to learn module structure with our inference algorithm is (in expectation) $n O(\log n)$, whereas the time to learn interaction coefficients is reduced from $O(n^2)$ to $O((\log n)^2)$.

We next applied our algorithm to real data from \cite{Bucci2016}, which investigated developing a bacteriotherapy for \textit{Clostridium difficile}, a pathogenic bacteria that causes serious diarrhea and is the most common cause of hospital acquired infection in the U.S. Five germ-free mice were colonized with a collection of 13 commensal (beneficial) bacterial species, termed the GnotoComplex microbiota, and monitored for 28 days (Figure \ref{fig:mdsine}A). Then, mice were infected with \textit{Clostridium difficile} and monitored for another 28 days. All mice developed diarrhea, but recovered within about a week, indicating that some combination of the 13 bacterial species protect against the pathogen (in a germ-free mouse, the infection causes death in 24-48 hours). Over the course of the experiment, 26 serial fecal samples per mouse were collected and interrogated via sequencing and qPCR to determine concentrations of the commensal microbes and the pathogen. We removed one species from our analysis, \textit{Clostridium hiranonis}, because it appeared to inconsistently colonize the mice, but otherwise used all data from the original study.

\begin{figure*}[htb!]\centering\vspace{-.1in}
\hspace{.2in}\begin{minipage}[T]{0.5\linewidth}
\includegraphics[width=1\textwidth]{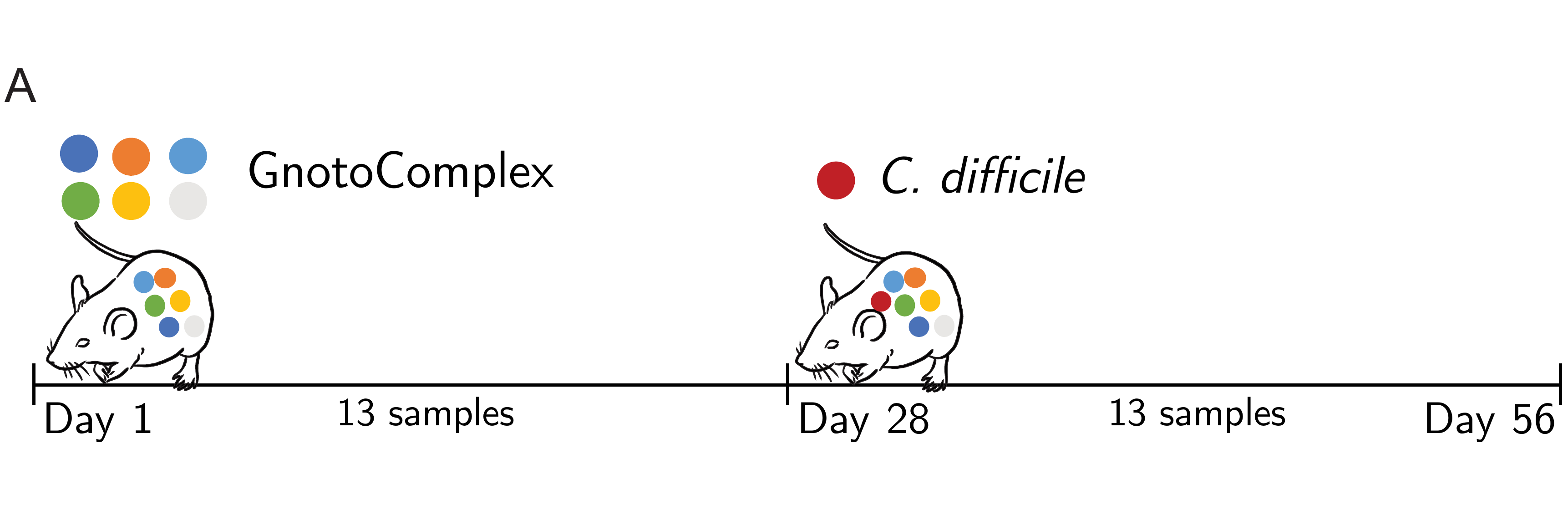}
\end{minipage}\hspace{.2in}
\begin{minipage}[T]{0.3\linewidth}\begin{tikzpicture}[>=stealth',scale=.8,every node/.style={scale=.8},auto,node distance=1cm,cir/.style={circle,draw},
 minimum size=25pt,inner sep=0pt,baseline=(current  bounding  box.center),red node/.style={circle,draw,color=gray,fill},squ/.style={draw},font=\tiny]
  \node[ellipse,draw] (a) [text width=38pt] {\textsf{\  \em C. scindens}};
  \node[ellipse,draw] (b) [right=25pt of a, text width=38pt] {\textsf{\em \ \ \  B. ovatus\\  \ P. distasonis}};
  \node[ellipse,draw]  (c) [below=25pt of b, text width=38pt]   {\textsf{\em  A. muciniphila\\  \ \ R. hominis}};
   	        \node[ellipse,draw]  (d) [left=25 pt of c, text width=38pt]   {\textsf{\em \ \  C. difficile\\  \  \ \ \  + rest}};
\node at (-1.15, .3)   (label) {\textsf{\footnotesize B}};
	      
      	\path[->](b)
	(a) edge  node[left] {$-$\textsf{2.1}}  (d) 
  	(b) edge node[below,yshift=2mm] {$-$\textsf{0.13}} (d)
  	(c) edge node[left,xshift=4mm,yshift=2mm] {$-$\textsf{0.05}} (d);
\end{tikzpicture}\end{minipage}\\
\includegraphics[width=.8\textwidth]{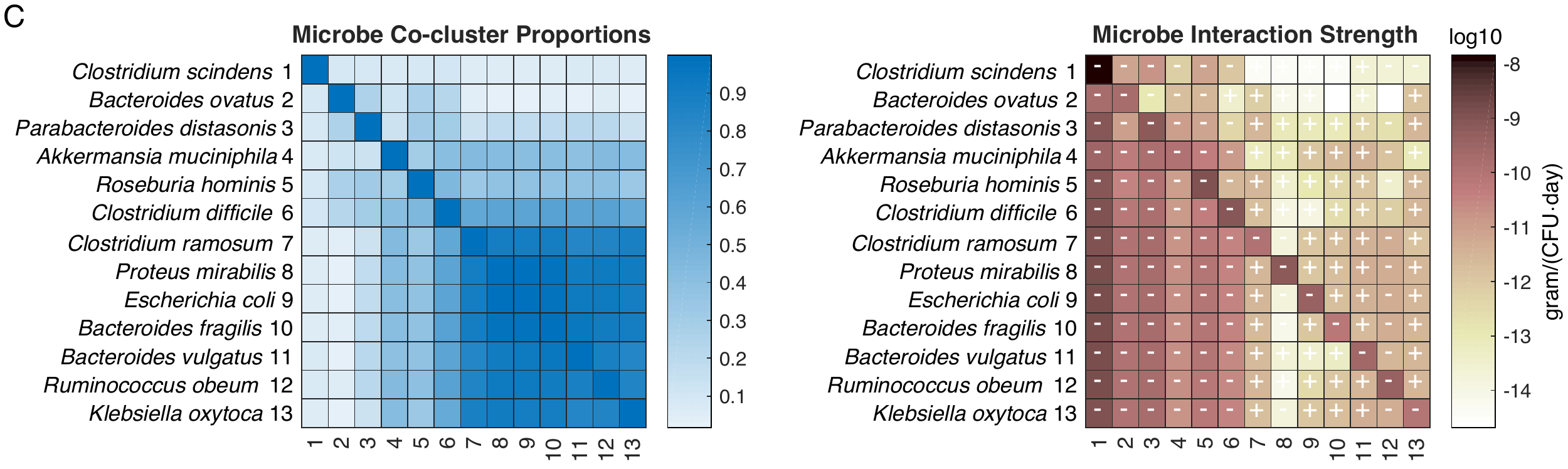}\vspace{-6pt}
\caption{ Inference applied to \emph{in vivo} experiments from \cite{Bucci2016}, illustrating the ability of interaction module learning to produce interpretable interaction structures that agree with biologically validated and plausible interactions. ({\bf A}) Experimental timeline (performed with 5 germ-free mice). GnotoComplex microbes, a defined collection of beneficial gut bacteria, is introduced on day one with \emph{Clostridium difficile} introduced on day 28. ({\bf B}) Module structure of a representative sample from the posterior with interaction strengths shown (interaction scale is $10^{-9}$). ({\bf C})    Co-cluster proportions illustrating the probability that two microbes appear in the same module and expected values for interaction coefficients, log10 scale with interaction signs illustrated.}
\label{fig:mdsine}
\end{figure*}
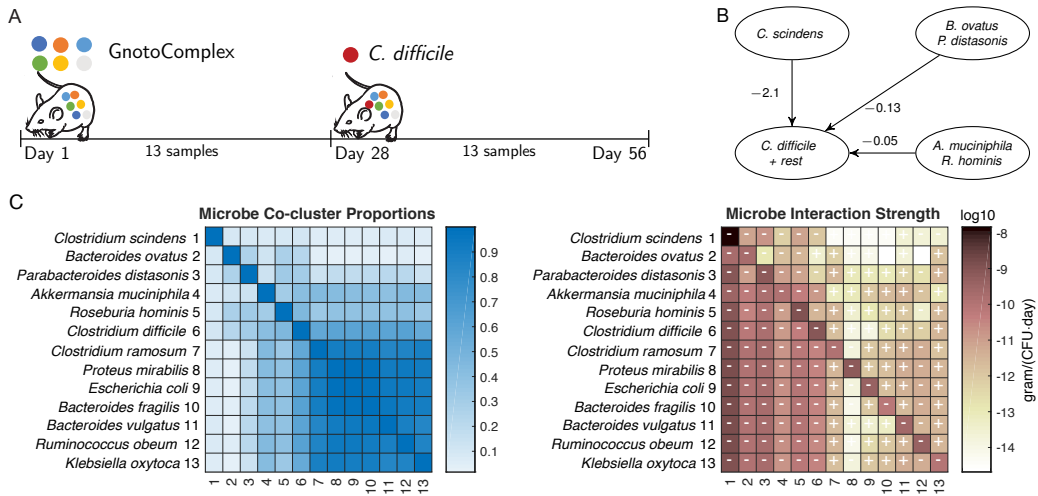

Figure \ref{fig:mdsine} shows the results of applying our model to the data from \cite{Bucci2016}. Our model found a median of 4 interaction modules (5,000 MCMC samples with 1,000 burnin). Seven microbes formed a large and consistent module, with the remaining six microbes aggregating into smaller modules. Figure \ref{fig:mdsine}B shows the module structure of a representative sample from the posterior. The module structure identifies groups of microbes that putatively inhibit the pathogen, and does so more clearly than in the original study, which presented a dense network of microbial interactions. The fine structure of this dense network is indeed still recapitulated in the posterior summary of interaction coefficients (Figure \ref{fig:mdsine}C), but our model also has the advantage of providing a compact module structure that is much easier to interpret biologically. Interestingly, the strongest interaction identified by our model (which the analysis from the original study detected relatively weakly), with \textit{Clostridium scindens} inhibiting the pathogen, is in fact the only biologically validated result in their study. Our analysis also discovered additional putative inhibitors of the pathogen, including the commensal \textit{Akkermansia munciniphila}. This microbe lives in the mucous layer in the gut, and has been associated positively with mucosal integrity in several studies (see e.g., \cite{belzer2017microbial}), and thus suggests an interesting and biologically plausible candidate for inclusion in a bacteriotherapy against the pathogen.

\section{Conclusions}
We have presented a Bayesian nonparametric model and associated inference algorithm for tackling key challenges in analyzing dynamics of the microbiome. Our method introduces several innovations, including a new type of modular dynamical systems model, uncertainty propagation throughout the model, and an efficient technique for approximating physically realistic non-negative dynamics. Applications of our method to simulated data show the ability to accurately identify the underlying dynamical system even with limited data. Application to real data highlights the ability of our model to infer compact, biologically interpretable representations that correctly find known relationships and suggest new, biologically plausible relationships. 

There are several areas for future work. Other Bayesian clustering approaches, which are more flexible than DPs, such as mixtures of finite mixtures \cite{miller2017mixture}, would be interesting to investigate as alternate priors for interaction modules. The gLV dynamical systems model has been widely used in microbial ecology, but has limitations including modeling only pairwise interactions and quadratic nonlinearities. Our inference method is quite flexible, and could readily accommodate other dynamical systems models, although nonlinearities in coefficients would cause difficulties (gLV is linear in the coefficients) in efficiency with our current algorithm. Another interesting avenue is using other forms of approximate inference to accelerate our algorithm, including approximate parallel MCMC and Variational Bayesian techniques. Incorporating prior biological knowledge, such as phylogenetic relationships between microbes, is another interesting area to investigate; because our model is fully Bayesian, incorporating prior knowledge is conceptually straight forward. Designing \emph{in vivo} experiments with sufficient richness to identify dynamical systems is a very important topic, and applying our model within a formal experimental design framework would thus be very interesting. On the application side, we plan to apply our model to additional bacteriotherapy design problems, which is an active and growing area of research. In this regard, our goal is to apply our model to upcoming human microbiome bacteriotherapy trials, which will measure the abundances of hundreds of gut commensal bacterial species per person.

\section*{Acknowledgements}
We thank the reviewers for their many helpful comments and suggestions. They greatly improved the final paper. This work was supported by NIH 5T32HL007627-33,  DARPA BRICS HR0011-15-C-0094 and the BWH Precision Medicine Initiative.

\bibliography{../../master}
\bibliographystyle{icml2018}

\clearpage

\twocolumn[
\icmltitle{Supplementary Material: Robust and Scalable Models of microbiome Dynamics}
\begin{icmlauthorlist}
\icmlauthor{Travis E.~Gibson}{}
\icmlauthor{Georg K.~Gerber}{}

\end{icmlauthorlist}
\vskip 0.3in
]
\appendix
\setcounter{equation}{4}

\section{Extended discussion regarding constraining dynamics}

We present an analysis of a naive model that directly constrains dynamics to be non-negative, to illustrate the issues this causes for the posterior distribution. Consider a dynamical process with latent state $\s x$, measurements $\s y$, and dynamical interaction coefficients $\s a$: 
\begin{equation}\label{gm:1}
\begin{tikzpicture}[>=stealth',auto,node distance=1.2cm,cir/.style={circle,draw},
 minimum size=.5cm,inner sep=0pt,baseline=(current  bounding  box.center),red node/.style={circle,draw,color=gray,fill},squ/.style={draw},font=\footnotesize,scale=0.75, every node/.style={transform shape}]

  \node[cir] (z31)  {$\s x_1$};
  \node[cir] (z32) [right of=z31] {$\s x_2$};
  \node[cir] (z33) [right of=z32] {$\s x_3$};
  \node(z3d) [right of=z33,node distance=1.0cm] {$\cdots$};
  \node[cir] (z3n) [right of=z3d,node distance=1.0cm] {$\s x_n$};
  \node[cir,minimum size=.5cm] (w3) [above left=0.2cm and .5cm of z31] {\color{black}$\s a$};
  \path[->]  
    (z31) edge (z32)
    (z32) edge (z33)
    (z33) edge (z3d)
    (z3d) edge (z3n)
    (w3) edge[out=0, in=135,looseness=.7] (z32)
    (w3) edge[out=0, in=135,looseness=.5] (z33)
    (w3) edge[out=0, in=135,looseness=.3] (z3n);
  \node[cir,fill=black!10] (y1)[below of=z31,node distance=1.2cm]   {$\s y_1$};
  \node[cir,fill=black!10] (y2) [right of=y1] {$\s y_2$};
  \node[cir,fill=black!10] (y3) [right of=y2] {$\s y_3$};
  \node(yd) [right of=y3,node distance=1.0cm] {$\cdots$};
  \node[cir,fill=black!10] (yn) [right of=yd,node distance=1.0cm] {$\s y_n$};

  \path[->]
  (z31) edge (y1)
    (z32) edge (y2)
     (z33) edge (y3)
      (z3n) edge (yn);
    \end{tikzpicture}\vspace{0pt}
    \end{equation}
  generated by the following 
\begin{equation}\label{eq:dyn1}
\begin{split}
\s x_{k+1,i}\mid \s x_{k}, {\s a} &\sim \mathtt{Normal}_{\geq 0}({{\s  a_i}^\mathsf{T}}f(\s x_k),\sigma_{\s x_i}^2) \\
\s y_{k,i}\mid \s x_{k,i} & \sim  \mathtt{Normal}_{\geq 0}(\s x_{k,i}, \sigma^2_{\s y_i}) \\
{\s a_i} & \sim \mathtt{Normal}(0,\sigma^2_{\s a_i}).
\end{split}
\end{equation}
The dynamics in \eqref{eq:dyn1} are precisely the dynamics one obtains via adding a  truncated normal measurement model to the discrete gLV dynamics presented in (1).\footnote{Note that this is the most direct way one can enforce a hard non-negativity constraint on the dynamics, and is indeed the first direction we took before realizing the challenges it imposes.} For ease of exposition let us assume for now that there is only 1 microbial species ($i=1$ and thus index $i$ can be dropped for this brief exposition) and all of the variance terms in \eqref{eq:dyn1} are equal to $\sigma^2$. Performing full Bayesian inference for $\p a$ requires constructing the posterior $p_{{\s a} \mid \s x}\propto p_{\s x \mid \s a} p_{\s a}$. Noting that the likelihood of $\p x$ satisfies the following proportionality $p_{{\s x} \mid \s a}  \propto \prod_{k} p_{{\s x_{k+1}} \mid \s a, \s x_{k}} $
and expanding this given our model in \eqref{eq:dyn1} we have
\begin{equation}\label{eq:like}
p_{{\s x} \mid \s a} (x \mid a) \propto  \prod_k \frac{\exp{-\frac{1}{2\sigma^2} (x_{k+1}-a^\mathsf T f(x_k)  )^2 }}{\sigma \sqrt{2\pi}  \left(\Phi(\infty) -{\Phi\left( -\frac{a^\mathsf T f(x_k)   }{\sigma}\right)}  \right) }
\end{equation}
where $\Phi$ is the Cumulative Distribution Function (CDF) for standard Normal distribution. 
Using the likelihood in \eqref{eq:like} and the prior for $\s a$ in \eqref{eq:dyn1}, the posterior of $\s a$ takes the form
\begin{multline*}
p_{{\s a} \mid \s x} (a \mid x) \\
\propto\prod_k \frac{\exp{-\frac{1}{2\sigma^2} (x_{k+1}-a^\mathsf T f(x_k)  )^2 }}{\sigma \sqrt{2\pi}  \left(\Phi(\infty) -{\Phi\left( -\frac{a^\mathsf T f(x_k)   }{\sigma}\right)}  \right) }
 \frac{\exp{-\frac{1}{2\sigma^2} { a ^\mathsf T a} }}{(\sigma^2 2 \pi)^{n_a/2}} 
\end{multline*}
where $n_a$ is the dimension of the column vector $a$. Having the variable $a$ appear in the normalization constant means we cannot directly Gibbs sample $\s a$, and also makes constructing an efficient proposal distribution in a Metropolis Hastings (MH) setting challenging too, as the proposal will in turn be affecting the scaling factor of the target distribution. A similar issue is encountered when trying to sample from the latent state $\s x \mid \s a, \s y$ (filtering).


%
%
%

\end{document}